# Automatic Target Recognition (ATR) from SAR Imaginary by Using Machine Learning Techniques

Umut Özkaya [1*]

[1]*Department of Electrical and Electronics Engineering/Konya Technical University, Turkey*
*(uozkaya@ktun.edu.tr) Email of the corresponding author*

*Abstract* – Automatic Target Recognition (ATR) in Synthetic aperture radar (SAR) images becomes a very challenging problem owing to containing high level noise. In this study, a machine learning-based method is proposed to detect different moving and stationary targets using SAR images. First Order Statistical (FOS) features were obtained from Fast Fourier Transform (FFT), Discrete Cosine Transform (DCT) and Discrete Wavelet Transform (DWT) on gray level SAR images. Gray Level Co-occurrence Matrix (GLCM), Gray Level Run Length Matrix (GLRLM) and Gray Level Size Zone Matrix (GLSZM) algorithms are also used. These features are provided as input for the training and testing stage Support Vector Machine (SVM) model with Gaussian kernels. 4-fold cross-validations were implemented in performance evaluation. Obtained results showed that GLCM + SVM algorithm is the best model with 95.26% accuracy. This proposed method shows that moving and stationary targets in MSTAR database could be recognized with high performance.

## I. INTRODUCTION

Synthetic aperture radar (SAR) is a device to obtain images in full time and actively [1]. SAR images are frequently used in reconnaissance, surveillance, target recognition and tracking for military application [2]. In recent years, recognition and detection of targets in SAR images has been increased to study day by day. Target recognition in SAR images a challenging task owing to including high level noise.

Automatic Target Recognition (ATR) process, which is planned to be performed on SAR images, has two stages. First of all, external factors, which are trees, cars, buildings, etc, reveal the false alarm situation. It is necessary to be omitted these from images. In the second stage, it performs feature extraction and classification algorithms [3].

Template matching technique, one of the traditional methods, is inadequate in target detection. The main reason for this is that there are changes in targets on SAR images due to the noise level [4]. In some studies, it has been tried to recognize the targets by obtaining local and global features [5]. Dong et al. obtained sparse representations of SAR images and recognize targets with different classifiers [6]. Zhou et al. carried out a multiscale feature fusion by performing canonical correlation analysis of sparse matrices. The fused features were classified for target recognition in SAR images [5]. Pan et al. analyzed the weighted sparse matrix from the SAR images with the classifier [7]. Liu et al. obtained features from SAR images. These were evaluated from two different classifiers and classification results were fused [8].

ATR operation on SAR images is observed as an ongoing problem. Raw SAR images include serious challenges. Obtaining SAR images is one of the most important problem. It also contains a high amount of noise in SAR images. Moving and Stationary Target Acquisition and Recognition (MSTAR) dataset is frequently used for ATR operations [9]. Novak et al. achieved 66.2% and 77.4% accuracy for 20-class and 10-class in the MSTAR data set respectively [10]. Martone et al. used k-means clustering algorithm for detection of moving targets in forested land [11]. Gorovyi and Sharapov achieved an accuracy rate of 90.7% with SVM on the MSTAR data set [12].

In this study, MSTAR data set was used. SAR images with $15^0$ degrees in the data set were used for both training and testing in classification algorithm. Two different strategies were followed for methodology. Fast Fourier Transform (FFT), Discrete Cosine Transform (DCT) and Discrete

Wavelet Transform (DWT) were applied to gray level SAR images. First Order Statistical (FOS) features were obtained from these transformation matrices. These features were classified with Support Vector Machine algorithm with Gaussian Kernels. Another strategy is to use gray level feature extraction methods. These methods are respectively Gray Level Co-occurrence Matrix (GLCM), Gray Level Run Length Matrix (GLRLM) and Gray Level Size Zone Matrix (GLSZM). The obtained features are processed in SVM with Gaussian kernels.

This paper is organized as follows. Section 2 includes materials and methods. Section 3 presents findings and discussion. Section 4 is conclusion part.

## II. MATERIALS AND METHODS

### A. Dataset Description

The data set used in the study was named as MSTAR. It was obtained by Defence Advanced Research Projects Agency (DARPA) and the American Air Force Research Laboratory (AFRL) [12]. SAR data were collected at different angles with the help of a radar operating in X band. In this study, analyses were performed on SAR images of 2S1, BRDM-2, BTR-60, D7, SLICY, T62, ZIL 131 and ZSU-23-4. Optical and SAR images of these classes are shown in Fig. 1.

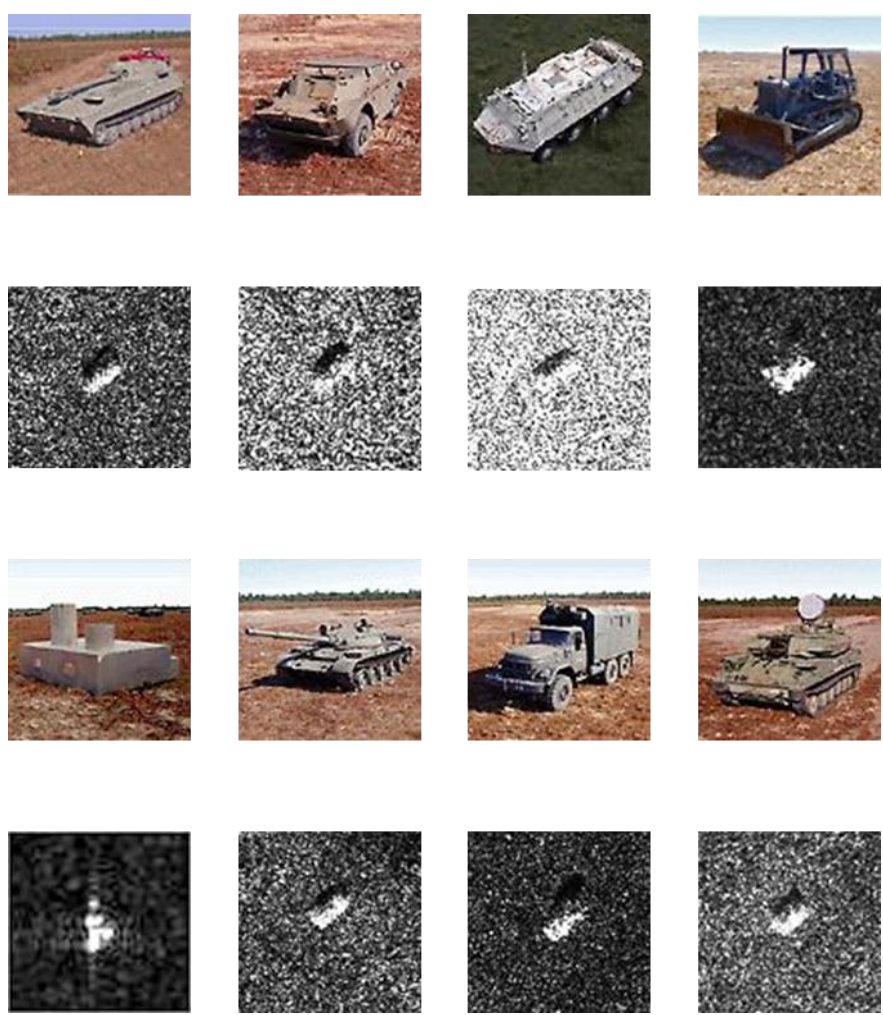

Fig. 1 The examples of 2S1, BRDM-2, BTR-60, D7, SLICY, T62, ZIL 131 and ZSU-23-4 with optical and SAR images

### B. Transformation Techniques

Discrete Fourier Transforms (DFT) converts a sequence in time space into an equivalent sequence in frequency space. FFT is a very efficient method based on DFT and requires much less computational load than DFT. FFT is widely used for frequency spectrum analysis in digital signal processing applications [13]. DCT is a common method in image compression. It is a method similar to DCT and it is a linear transformation. [14]. DWT is a filter bank that separates the image into frequency sub-bands [15]. Where horizontal details refer to horizontal high frequencies, vertical details to vertical high frequencies, diagonal details to high frequencies in both directions. The features were obtained by using LL, LH, HL and HH coefficients from DWT. Six features were extracted from each coefficient as mean, variance, kurtosis, skewness, entropy and energy.

*C. Gray Level Feature Extraction Techniques*

GLCM is a feature set consisting of second order statistical features. GLCM is created by considering the relationships between the pixels of an image from different angles. Covariance matrices obtained from an image can be expressed as P = [p (i,j|d, Θ)]. Where i. pixel frequency properties and j are used to evaluate frequency features of neighbouring pixels with reference to d distance and Θ direction. GLCM features can be defined as angular second moment, contrast, correlation, sum of squares of variance, inverse moment of difference, total mean, total variance, total entropy, entropy, difference of variance, entropy difference, correlation information criterion 1, correlation information criterion 2, autocorrelation, dissimilarity, cluster tone, cluster prominence, maximum probability, and inverse difference [16].

GLRLM is a method of extracting high level texture features. Where G represents the number of gray levels, R is the longest run W the number of pixels in the image. GLRLM matrix is in G×R dimension. Each p (i,j| θ) element gives the number of occurrences in the θ direction at the i gray level and j run length. Seven different statistical features are obtained as short run emphasis, long run emphasis, gray level irregularity, run length irregularity, running percentage, low gray level running emphasis and high gray level running emphasis [16]. GLSZM is a feature extraction technique that has added two new features to GLCM method as size and density of a texture in the image [17].

*D. Support Vector Machine*

SVM is a method of achieving high performance in many applications. SVM is based on two key views. . The first idea is to map high dimensional space in a nonlinear method. It makes using of class classifiers in this new space. The second view is to find appropriate hyperplane that separates the data by a large margin. This plane separates the data as well as possible between an infinite numbers of planes [18].

*E. Proposed Frameworks*

This study includes two different strategies to classify SAR images. In the first strategy, some transformation techniques were applied on SAR images. Then, FOS features were extracted. The second strategy based on gray level features extraction. At last, SVM was performed on these features to classify for ATR. Proposed frameworks are given in Fig. 2.

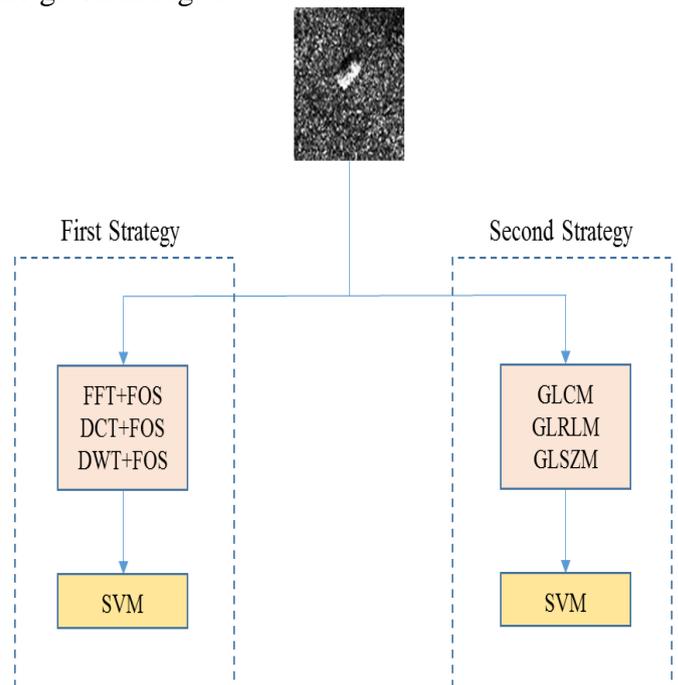

Fig. 2 Proposed Frameworks

III. RESULTS AND DISCUSSION

In this study, transformation techniques and gray level feature extraction algorithms were used. Classification process was carried out with SVM, which is one of the machine learning methods. Training and testing of SVM classifier was carried out using SAR images in $15^0$ degree from the MSTAR database. SAR image in $15^0$ degree numbers for each class are given in Table 1.

Table 1. Number of SAR Images for Each Classes

| Class | Number of Images |
|---|---|
| 2S1 | 274 |
| BRDM-2 | 274 |
| BTR-60 | 195 |
| D7 | 274 |
| SLICY | 274 |
| T62 | 273 |
| ZIL 131 | 274 |
| ZSU-23-4 | 274 |
| Total | 2112 |

Six different classification metrics were used to evaluate the proposed feature extraction and performance of SVM model. These metrics are Accuracy (ACC), Precision (SEN), Specificity (SPE), F1-Score, Matthews Correlation Coefficient (MCC). They are given in Eq. 1-6.

$$Accuracy = (TP+TN)/(TP+FN+TN+FP) \quad (1)$$

$$Sensitivity = TP/(TP+FN) \quad (2)$$

$$Specificity = TN/(TN+FP) \quad (3)$$

$$Precision = TP/(TP+FP) \quad (4)$$

$$F1-Score = (2 \times TP)/(2 \times TP + FN + FP) \quad (5)$$

$$MCC = \frac{TP \times TN - FP \times FN}{\sqrt{(TP+FP)(TP+FN)(TN+FP)(TN+FN)}} \quad (6)$$

Where TP is True Positive, TN is True Negative, FP is False Positive and FN is False Negative.

Mean pf metric performances and standard deviations of the proposed methods are given in Table 2.

Table 2. Performance of Proposed Frameworks

| Methods | Evaluation Metrics (%) | | | | | |
|---|---|---|---|---|---|---|
| | SEN | SPE | ACC | PRE | F1-Score | MCC |
| FOS+SVM | 74.41±0.72 | 96.64±0.10 | 76.60±0.71 | 75.77±1.31 | 73.16±0.95 | 70.97±0.82 |
| FFT+FOS+SVM | 73.33±1.44 | 96.31±0.22 | 74.24±1.59 | 74.16±2.05 | 73.21±1.64 | 69.86±1.89 |
| DCT+FOS+SVM | 69.36±0.95 | 95.85±0.14 | 71.02±1.01 | 69.99±0.74 | 68.87±0.93 | 65.25±1.02 |
| DWT+FOS+SVM | 50.87±1.88 | 93.13±0.25 | 52.13±1.75 | 53.52±2.29 | 49.83±2.00 | 44.64±2.25 |
| **GLCM+SVM** | **95.02±1.34** | **99.32±0.16** | **95.26±1.14** | **95.08±1.14** | **95.00±1.26** | **94.36±1.41** |
| GLRLM+SVM | 66.05±1.34 | 95.27±0.18 | 67.04±1.26 | 67.32±1.32 | 65.69±1.46 | 61.62±1.54 |
| GLSZM+SVM | 88.74±1.64 | 98.43±0.26 | 89.06±1.84 | 89.11±1.66 | 88.72±1.94 | 87.30±1.54 |

All metrics were computed in means and standard deviation. In Table 2, GLCM + SVM algorithm achieved the highest performance with 95.02 ± 1.34% SEN, 99.32 ± 0.16% SPE, 95.26 ± 1.14% ACC, 95.08 ± 1.14% PRE, 95.00 ± 1.26% F1-Score and 94.36 ± 1.41% MCC. The lowest performance belongs to the DWT + FOS + SVM model. The performance of this model is 50.87 ± 1.88% SEN, 93.13 ± 0.25% SPE, 52.13 ± 1.75% ACC, 53.52 ± 2.29% PRE, 49.83 ± 2.00% F1-Score and 44.64 ± 2.25% MCC. Table 3 consists of GLCM+SVM metric performance for each folds.

Table 3. GLCM+SVM Performance for Each Fold

| Evaluation Metrics (%) | 4-fold Cross Validation | | | |
|---|---|---|---|---|
| | Fold 1 | Fold 2 | Fold 3 | Fold 4 |
| SEN | 93.01 | 95.68 | 95.50 | **95.87** |
| SPE | 99.08 | 99.40 | 99.37 | **99.43** |
| ACC | 93.56 | 95.83 | 95.64 | **96.02** |
| PRE | 93.37 | 95.59 | 95.57 | **95.79** |
| F1-Score | 93.10 | 95.60 | 95.52 | **95.78** |
| MCC | 92.25 | 95.03 | 94.91 | **95.25** |

Table 3 shows the cross validation performance of GLCM+SVM framework. It is seen that the highest performance is obtained Fold 4 using as validation. These metric values are observed as 95.87% SEN, 99.43% SPE, 96.02% ACC, 95.79% PRE, 95.78% F1-Score and 95.25% MCC. The lowest performance belongs to Fold 1 data for validation is with 93.01% SEN, 99.08% SPE, 93.56% ACC, 93.37% PRE, 93.10% F1-Score and 92.25% MCC.

IV. CONCLUSION

In this study, ATR framework from SAR images based on machine learning methods was proposed. SVM algorithm with Gaussian kernels is used after obtaining features with two different strategies from gray level SAR images. It can be seen that GLCM+SVM model is quite successful. It is obvious that gray level features extraction methods show higher performance compared to transformation and FOS features performed on the MSTAR dataset. In the scope of proposed framework, it has been proven that moving or stationary targets in SAR images can be detected successfully.